%% file: root.tex
\documentclass[letterpaper, 10 pt, conference]{ieeeconf}  

\IEEEoverridecommandlockouts                              

\usepackage{graphicx} 
\usepackage{epsfig} 
\usepackage{times} 
\usepackage{amsmath} 
\usepackage{amssymb}  
\usepackage{url}
\usepackage{cite}
\usepackage{enumerate}
\usepackage{bm}
\usepackage{cases}
\usepackage{mathtools}
\usepackage[caption=false, font=footnotesize]{subfig}
\usepackage{algorithmicx}
\usepackage{algorithm}
\usepackage{algpseudocode}
\usepackage{booktabs, ragged2e}
\usepackage{tabularx}
\usepackage{xcolor}
\usepackage[para]{threeparttable}
\makeatletter
\let\NAT@parse\undefined
\makeatother
\usepackage[hidelinks]{hyperref} 
\usepackage[acronym]{glossaries}
\makeglossaries

\usepackage{siunitx}

\newacronym{EKF}{EKF}{Extended Kalman Filter}
\newacronym{iEKF}{iEKF}{Iterated Extended Kalman Filter}
\newacronym{MAP}{MAP}{maximum a-posteriori estimate}
\newacronym{AKF}{AKF}{Adaptive Kalman Filter}
\newacronym{LO}{LO}{LiDAR Odometry}
\newacronym{LIO}{LIO}{LiDAR-Inertial Odometry}
\newacronym{VIO}{VIO}{Visual-Inertial Odometry}
\newacronym{VO}{VO}{Visual Odometry}
\newacronym{IMU}{IMU}{Inertial Measurement Unit}
\newacronym{SLAM}{SLAM}{Simultaneous Localization and Mapping}
\newacronym{SVD}{SVD}{Singular Value Decomposition}
\newacronym{FoV}{FoV}{Field of view}
\newacronym{RTE}{RTE}{Relative Translation Error}
\newacronym{RMSE}{RMSE}{Root Mean Square Error}
\newacronym{ATE}{ATE}{Absolute Trajectory Error}

\title{\LARGE \bf
AKF-LIO: LiDAR-Inertial Odometry with Gaussian Map by Adaptive Kalman Filter
}

\author{Xupeng~Xie$^{1,2}$, Ruoyu~Geng$^{3}$, Jun~Ma$^{1,3}$, Boyu~Zhou$^{4,\dagger}$ 
\thanks{$^{\dag}$Corresponding Author} 
\thanks{$^{1}$The Hong Kong University of Science and Technology (Guangzhou).} 
\thanks{$^{2}$International Digital Economy Academy.}%
\thanks{$^{3}$The Hong Kong University of Science and Technology.}%
\thanks{$^{4}$Southern University of Science and Technology.}%
\thanks{{Email: {xxieak@connect.hkust-gz.edu.cn}, rgengaa@connect.ust.hk,}}\thanks{{\qquad \quad jun.ma@ust.hk, zhouby@sustech.edu.cn}}
\thanks{This work is supported by Shenzhen Science and Technology Program (NO. KJZD20240903103210014).}
}

\begin{document}

\maketitle
\thispagestyle{empty}
\pagestyle{empty}

\begin{abstract}

Existing LiDAR-Inertial Odometry (LIO) systems typically use sensor-specific or environment-dependent measurement covariances during state estimation, leading to laborious parameter tuning and suboptimal performance in challenging conditions (e.g., sensor degeneracy and noisy observations). Therefore, we propose an \acrfull{AKF} framework that dynamically estimates time-varying noise covariances of LiDAR and \acrfull{IMU} measurements, enabling context-aware confidence weighting between sensors.
During LiDAR degeneracy, the system prioritizes \acrshort{IMU} data while suppressing contributions from unreliable inputs like moving objects or noisy point clouds.
Furthermore, a compact Gaussian-based map representation is introduced to model environmental planarity and spatial noise. A correlated registration strategy ensures accurate plane normal estimation via pseudo-merge, even in unstructured environments like forests. 
Extensive experiments validate the robustness of the proposed system across diverse environments, including dynamic scenes and geometrically degraded scenarios. Our method achieves reliable localization results across all MARS-LVIG sequences and ranks 8th on the KITTI Odometry Benchmark. The code will be released at \href{https://github.com/xpxie/AKF-LIO.git}{https://github.com/xpxie/AKF-LIO.git}.

\end{abstract}

\input{sections/introduction.tex}

\input{sections/related_work.tex}

\input{sections/methodology.tex}

\input{sections/experiment.tex}











\bibliographystyle{IEEEtran}
\bibliography{IEEEabrv,ref}

\end{document}

%% file: sections/introduction.tex
\section{Introduction}

\acrfull{SLAM} refers to the task of concurrently estimating a robot's state (e.g., position and orientation) and reconstructing a map of the environment by probabilistically fusing multi-modal sensor data. Sensors like LiDAR, camera and \acrfull{IMU} are utilized in multiple ways to complement each other. \acrfull{LIO} has emerged as a dominant approach among these methods, leveraging high-precision LiDAR ranging information and high-frequency \acrshort{IMU} measurements for robust localization and dense mapping.

Existing \acrshort{LIO} systems still face two critical challenges despite their widespread adoption. First, they lack a generalized framework for online noise covariance estimation. Most methods rely on static sensor noise models \cite{xu2022fast, bai2022faster} or complex sensor modeling \cite{yuan2022efficient}, leading to over-constrained or under-constrained optimization during sensor degradation or time-varying sensor noise. This results in unreliable state estimation, particularly when high confidence is assigned to measurements from LiDAR-degenerated directions or dynamic objects. 
Second, geometric fidelity in LiDAR maps is compromised by inadequate map representations. Conventional point clouds tend to misclassify planar surfaces as edges due to sparsity and irregularity of LiDAR data \cite{xu2022fast, bai2022faster, chen2023direct}. While advanced representations like surfel \cite{behley2018efficient} and Gaussian primitives \cite{yuan2022efficient, chen2024ig, ji2024lio} improve geometric modeling, their resolution remains bounded by fixed voxel sizes.

\begin{figure}[t]
  \vspace{0.5em}
  \centering
  \includegraphics[width=3.4in]{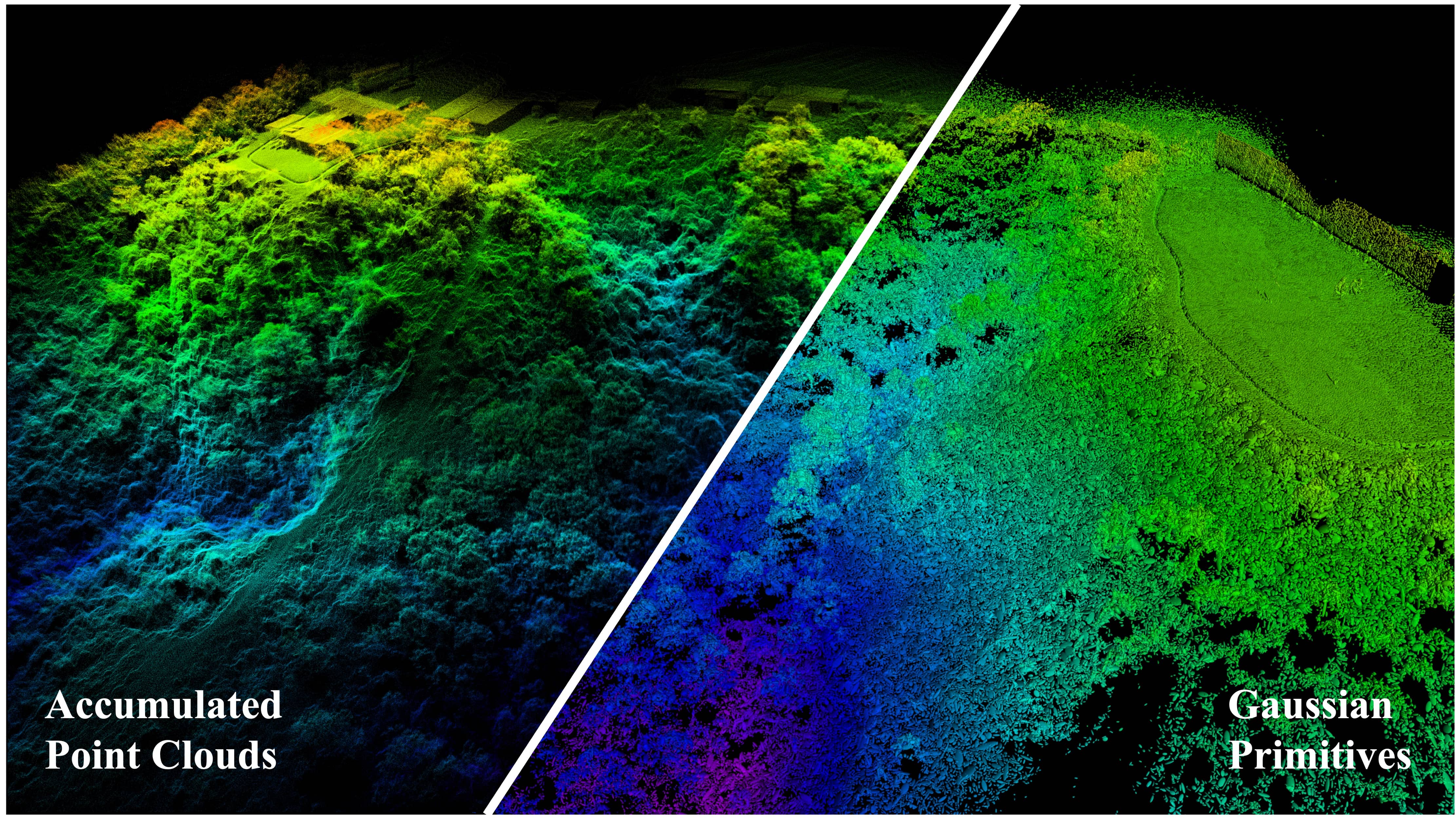}
  \vspace{-1.5em}
  \caption{The mapping result of Featureless\_GNSS03 sequence in MARS-LVIG dataset \cite{li2024mars}. Left: point cloud map, right: Gaussian map, both colored by height.}
  \label{fig::mars03_g}
  \vspace{-1.5em}
\end{figure}

To address these issues, we propose an \acrfull{AKF}-based \acrshort{LIO} framework with multi-scale Gaussian map, assuming environments consist of planar structures. The \acrshort{AKF} dynamically estimates process and measurement noise covariances via innovation and residual terms \cite{akhlaghi2017adaptive}. While \acrshort{AKF} is primarily applied to INS/GPS systems \cite{mohamed1999adaptive, sasiadek2000fuzzy}, adapting it to LiDAR is non-trivial: verifying whether a LiDAR point belongs to a static planar surface remains challenging due to sparse and noisy neighboring points. 
In contrast, Gaussian distribution inherently provides continuity and outlier robustness.
Therefore, our framework estimates the measurement covariance for each Gaussian map primitive by integrating its current and previous residuals. A fine-to-coarse voxelized Gaussian map preserves geometric details for small structures while aggregating large planar surfaces.
Furthermore, we introduce a Gaussian-based pseudo-merge strategy to ensure accurate plane normal estimation during registration, even in unstructured environments like forests in  Fig. \ref{fig::mars03_g}. This approach guarantees that the LiDAR point to be registered lies within high-confidence regions of the pseudo-merged Gaussian.
The contributions of this paper include:
\begin{enumerate}[1)]
  \item We propose an online noise covariance estimation module for LiDAR and IMU measurements by \acrshort{AKF}, enhancing robustness and generalizability across diverse sensor configurations and environments.
  \item We propose a multi-scale Gaussian map representation constructed in a fine-to-coarse manner, paired with a correlated registration strategy, to achieve high-fidelity geometric modeling and stable plane normal estimation.
  \item We verify the accuracy and robustness of our proposed method through various experiments across diverse environments, especially in dynamic and geometrically degenerated environments.
\end{enumerate}

%% file: sections/related_work.tex
\section{Related work}

\subsection{Uncertainty in LiDAR(-Inertial) Odomerty}
Robust uncertainty modeling is essential for ensuring stability and reliability in \acrshort{LO}/\acrshort{LIO} systems, of which error sources are categorized into sensor noise and pose estimation degeneracy \cite{jiao2021robust}. 
PUMA-LIO \cite{jiang2022lidar} introduces a sampling-based method to estimate noise along the plane normal direction and surface roughness. LOG-LIO2 \cite{huang2024log} further proposes a comprehensive point uncertainty model incorporating range, bearing, incident angle and surface roughness. VoxelMap \cite{yuan2022efficient} extends these efforts by jointly addressing sensor noise and pose uncertainty through point-wise uncertainty aggregation into plane uncertainty.

Our method employs \acrshort{AKF} to dynamically estimate both measurement and process covariances, enhancing resilience against sensor degradation and environmental changes. 
Unlike VoxelMap \cite{yuan2022efficient} which relies on complex physical sensor models, our method reduces reliance on prior statistical assumptions through self-adaptive covariance updates. While VoxelMap propagates pose uncertainty to all LiDAR points, we explicitly quantify constraint uncertainty per map element, enabling discrimination between well-constrained and under-constrained regions during LiDAR degeneracy. Additionally, our direct uncertainty updates on map primitives achieve faster adaptation to LiDAR degeneracy compared to VoxelMap's point-wise uncertainty propagation.

\subsection{Map Representation in LiDAR(-Inertial) Odomerty}
\label{sec:gaussian related work}

\begin{figure}[t]
  \centering

  \subfloat[Uniform voxel]{\includegraphics[width=0.32\linewidth]{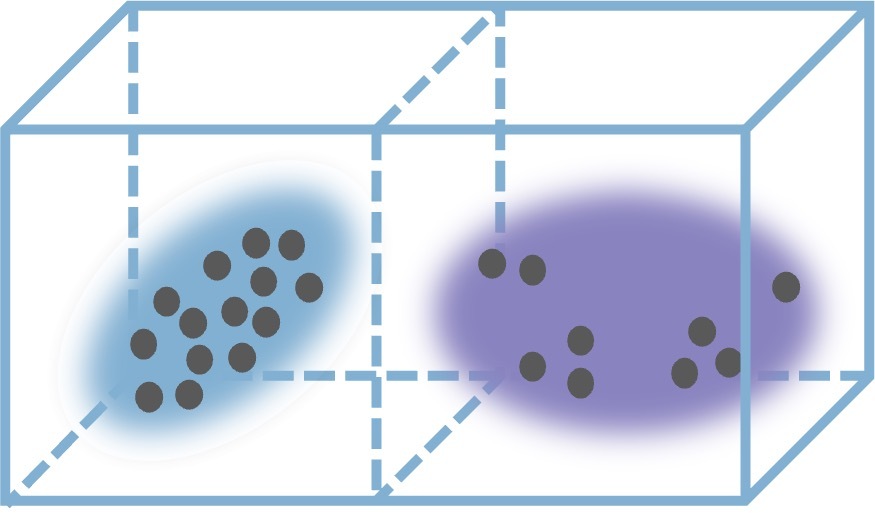}
  \label{fig:g map_a}}
  \subfloat[Octree]{\includegraphics[width=0.32\linewidth]{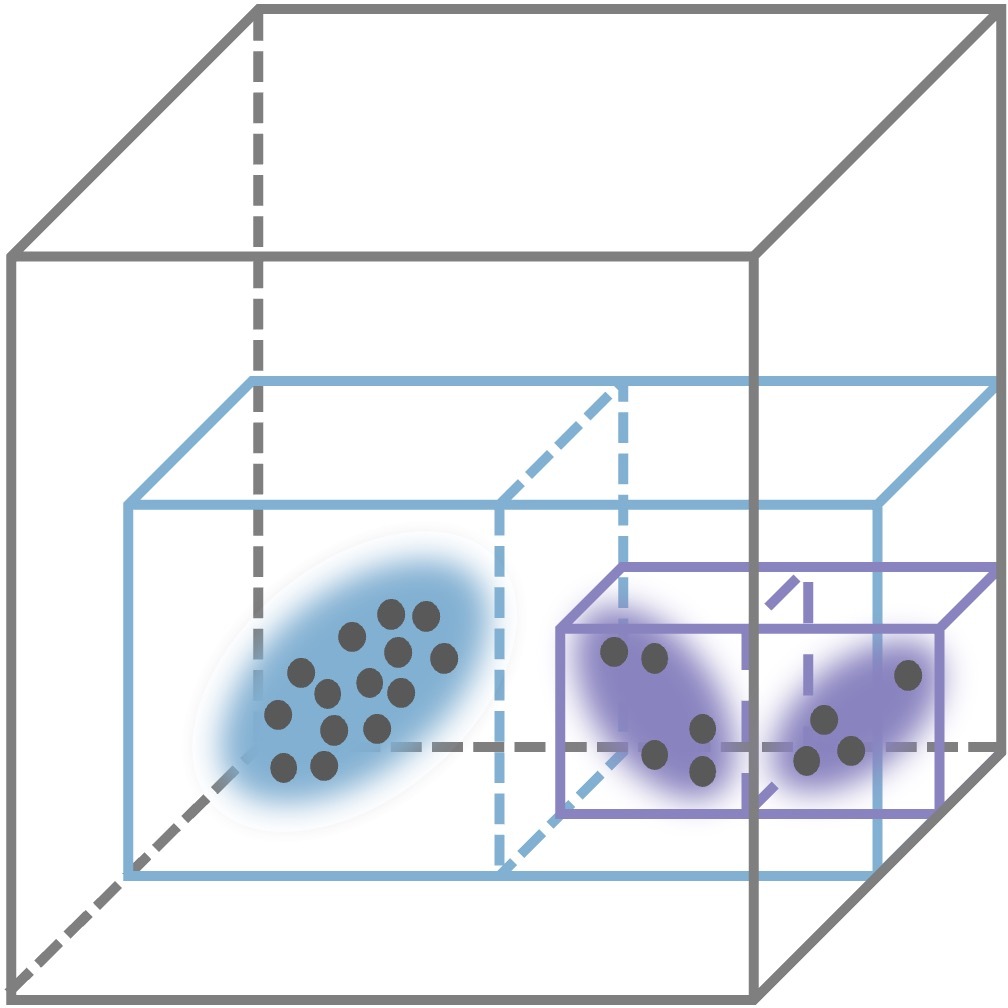}
  \label{fig:g map_b}}
  \subfloat[Adaptive size (ours)]{\includegraphics[width=0.32\linewidth]{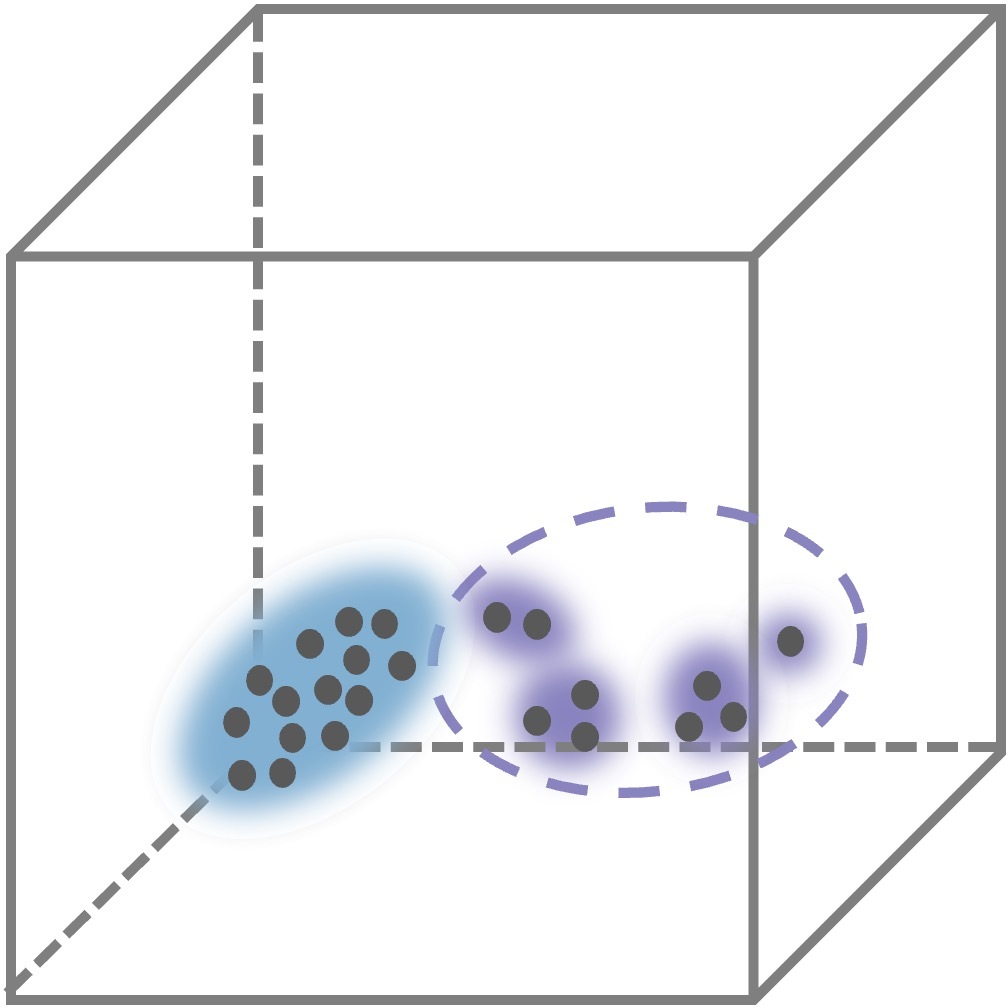}
  \label{fig:g map_c}}
 
  \caption{Different Gaussian-based map representations and their geometric resolution limits. (a) Uniform voxel-based Gaussian \cite{yokozuka2021litamin2, ji2024lio, chen2024ig} is limited by voxel size. (b) Octree-based Gaussian \cite{yuan2022efficient} is limited by tree depth. (c) Gaussian with adaptive size (ours) is not limited by voxel size indicated by the small Gaussians inside the purple circle.}
  \label{fig:gaussian map}
  \vspace{-0.5cm}
\end{figure}

Point cloud is a widely adopted map representation in \acrshort{LO}/\acrshort{LIO}, but their computational efficiency is compromised by the massive volume of LiDAR points.
To address this problem, FAST-LIO2 \cite{xu2021fast} implements a lazy update strategy via a sparse ikd-Tree for incremental maintenance, while Faster-LIO \cite{bai2022faster} enhances parallelization efficiency through voxel-based partitioning.
Alternatively, Gaussian distribution has emerged as a compact representation for modeling geometric primitives (e.g., planes or edges) using \acrfull{SVD}. LiTAMIN2 \cite{yokozuka2021litamin2}, LIO-GVM \cite{ji2024lio}, and iG-LIO \cite{chen2024ig} utilize fixed-size per-voxel Gaussians, but their geometric resolution remains bounded by predefined voxel size as illustrated in Fig. \ref{fig:g map_a}. VoxelMap \cite{yuan2022efficient} enhances this by adopting an octree structure for adaptive voxel sizing, but it requires to store raw LiDAR points for dynamic voxel subdivision and its minimum resolution is limited by the tree depth as shown in Fig. \ref{fig:g map_b}. VoxelMap++ \cite{wu2023voxelmap++} further improves accuracy and efficiency via a union-find-based plane merging strategy to construct larger planes, but it retains voxel-level resolution limits.

Our method introduces a fine-to-coarse Gaussian map representation, assuming environments comprise multi-scale planar structures. Unlike VoxelMap++ \cite{wu2023voxelmap++} which confines resolution to its minimum voxel size, our approach permits multiple Gaussians within a single voxel as demonstrated in Fig. \ref{fig:g map_c}. This design preserves point-wise geometric details while aggregating large planar regions, surpassing the voxel-level resolution constraints of existing methods.

%% file: sections/methodology.tex
\section{System Overview}
\begin{figure*}[t]
  \vspace{0.5em}
  \centering
  \includegraphics[width=6.8in]{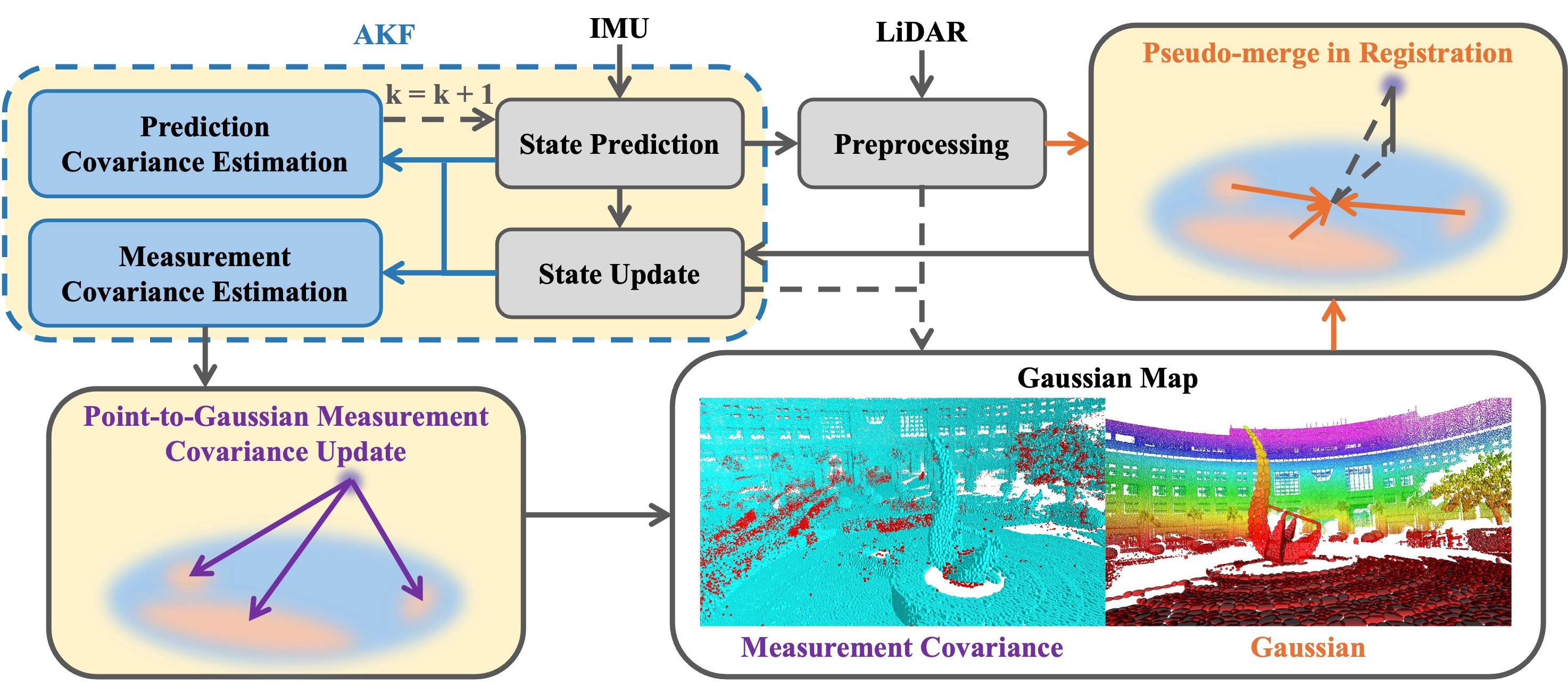}
  \caption{Framework overview of AKF-LIO. \acrshort{AKF} estimates state by fusing \acrshort{IMU} and LiDAR measurements with self-adjustment of prediction and measurement covariances for robustness to sensor degeneracy and environmental changes. The Gaussian map (hkust\_campus\_02 sequence in R3LIVE dataset\cite{lin2022r}) maintains planar structures represented by Gaussian distribution (colored by height) and their measurement covariances (red: high covariance, blue: low covariance) which are utilized in the registration module with pseudo-merge strategy.}
  \label{fig::framework}
\end{figure*}

As illustrated in Fig. \ref{fig::framework}, the proposed system consists of three core components: an \acrshort{AKF}, a voxel-based Gaussian map and a registration module. Synchronized \acrshort{IMU} measurements are used for state prediction with estimated prediction covariance, enabling subsequent LiDAR scan motion compensation and state update. The undistorted point cloud after downsampling is aligned to the Gaussian map via pseudo-merge strategy weighted by dynamically estimated measurement covariances for state update. The refined state, innovation, and residual terms are leveraged to update prediction and measurement covariances to adaptively tune the filter's parameters. Concurrently, the current LiDAR scan and its associated measurement covariances are integrated into the Gaussian map using the refined state. 
\begin{table}[t]
\caption{State Notation}
\begin{tabularx}{\columnwidth}{@{\extracolsep{\fill} } l l}
\toprule

$\mathbf{x}$, $\widehat{\mathbf{x}}$, $\bar{\mathbf{x}}$ & Ground truth, prediction and update of state $x$. \\

$\widetilde{\mathbf{x}}$ & Error between ground truth state $\mathbf{x}$ and prediction $\widehat{\mathbf{x}}$. \\

$\delta {\mathbf{x}}$ & Error between update state $\bar{\mathbf{x}}$ and its prediction $\widehat{\mathbf{x}}$. \\

$\mathbf{x}_i$ & State $\mathbf{x}$ at the $i$th IMU sampling time.\\

$\mathbf{x}_k$ & State $\mathbf{x}$ at the $k$th LiDAR scan end time. \\

$\mathbf{x}^\kappa$ & State $\mathbf{x}$ at the $\kappa$th iteration. \\

${}^W(\cdot)$, ${}^L(\cdot)$, ${}^I(\cdot)$ & Variables in the World, LiDAR and IMU frame. \\

\bottomrule
\end{tabularx}
\label{table:state notations}
\end{table}

\begin{table}[t]
\centering
\caption{Variable Definition}
\begin{tabularx}{\columnwidth}{@{\extracolsep{\fill} } l l}
  \toprule
  
  ${^W\mathbf R}_I$, $^W\mathbf p_I$ & Rotation and translation of \acrshort{IMU} w.r.t. World. \\
  
  $^W\mathbf v_I$ & Velocity of \acrshort{IMU} w.r.t. World. \\
  
  $\mathbf b_{\bm \omega}$, $\mathbf b_{\mathbf a}$ & Bias vectors of \acrshort{IMU} gyroscope and accelerometer. \\
  
  $^W\mathbf g$ & Gravity vector w.r.t. World. \\
  
  $\bm \omega$, ${\mathbf a}$ & \acrshort{IMU} gyroscope and accelerometer measurements. \\

  $\mathbf n_{\bm \omega }$, $\mathbf n_{\mathbf a}$ & Measurement noise of $\bm \omega$ and ${\mathbf a}$. \\

  $\mathbf n_{\mathbf b\bm \omega }$, $\mathbf n_{\mathbf b\mathbf a}$ & Noise of  $\mathbf b_{\bm \omega}$ and $\mathbf b_{\mathbf a}$.  \\
  \bottomrule
\end{tabularx}
\label{table:variable definition}
\end{table}

\section{Adaptive Kalman Filter}
\acrshort{AKF} in our system is built upon the \acrfull{iEKF} in FAST-LIO2 \cite{xu2022fast} with additional prediction and measurement covariance estimation modules to replace the original fixed \acrshort{IMU} noise covariance and LiDAR constraint uncertainty used in FAST-LIO2. The critical notations of the state are summarized in Table \ref{table:state notations}.

\subsection{Problem Definition}
\label{sec:problem definition}

The state $\mathbf{x}$ is defined as follows:
\begin{equation} 
\begin{aligned} 
  \mathbf x &\triangleq \!\begin{bmatrix}^W{\mathbf R}_I^T\!&\!^W\mathbf p_I^T\!\!&\!^W\mathbf v_I^T\!&\!\mathbf b_{\bm \omega }^T\!&\!\!\mathbf b_{\mathbf a}^T\!\!&\!\!^W\mathbf g^T \end{bmatrix}^T\!\in \!\! \mathcal M,\\ 
  \mathcal {M} &\triangleq \text{SO}(3) \times \mathbb {R}^{15},\ \dim (\mathcal {M}) = 18, \\
  \mathbf u &\triangleq \begin{bmatrix}\bm \omega ^T_m & {\mathbf a}^T_m\end{bmatrix}^T,\ \mathbf w \triangleq \begin{bmatrix}\mathbf n_{\bm \omega }^T&\mathbf n_{\mathbf a}^T&\mathbf n_{\mathbf b\bm \omega }^T&\mathbf n_{\mathbf b\mathbf a}^T\end{bmatrix}^T, \\ 
  \end{aligned} 
\end{equation}
and the involved variables are defined in Table \ref{table:variable definition}.
\acrshort{iEKF} aims to iteratively refine the error state $\widetilde{\mathbf{x}}$ for linearization accuracy and computational efficiency. 
Under the assumption that the environment consists of planar structures, the point-to-plane residual is used for the registration between the $j$th LiDAR point ${}^{L} \mathbf p_{j}$ with noise ${}^{L}{\mathbf n}_j$ and the map as defined in (\ref{eq:residual}). Estimating $\widetilde{\mathbf x}_{k}^\kappa$ can be formulated as an \acrshort{MAP} problem by regarding $\mathbf{x}_k \boxminus \widehat{\mathbf{x}}_k$ as its prior and combining the point-to-plane residual as the measurement model:
\begin{equation} 
\begin{aligned}
  \min _{\widetilde{\mathbf x}_{k}^\kappa } \left(\Vert \mathbf x_k \boxminus \widehat{\mathbf x}_k \Vert ^2_{ \widehat{\mathbf P}_k} + \sum \limits _{j=1}^{m} \Vert \mathbf z_{k,j}^\kappa + \mathbf H_{k,j}^\kappa \widetilde{\mathbf x}_{k}^\kappa \Vert ^2_{\widehat{\mathbf R}_{k,j}} \right),
\end{aligned}
\end{equation}
where $\mathbf H_{k,j}^\kappa$ is the Jacobian of the residual w.r.t. $\widetilde{\mathbf x}_{k}^\kappa$, $\widehat{\mathbf P}_k$ is the \acrshort{IMU} propagated state covariance, $\widehat{\mathbf R}_{k,j}$ is the uncertainty of the residual, and $\boxplus$ / $\boxminus$ refers to the mapping between the manifold $\mathcal M$ and its local tangent space $\mathbb {R}^{18}$ \cite{hertzberg2013integrating, xu2021fast}.

\subsection{State Prediction}
From $k-1$th LiDAR scan end to $k$th LiDAR scan end time, the state is propagated using the synchronized \acrshort{IMU} measurements. Combining the state transition function  $\mathbf f (\mathbf x, \mathbf u, \mathbf w)$ defined in FAST-LIO2 \cite{xu2022fast}, state is propagated as follows using the $i$th IMU measurement $\mathbf u_i$:
\begin{equation} 
\begin{aligned} 
  \widehat{ \mathbf x}_{i+1} &= \widehat{ \mathbf x}_{i} \boxplus \left(\Delta t \mathbf f (\widehat{\mathbf x}_i, \mathbf u_i, \mathbf 0) \right), \ \widehat{\mathbf x}_{0} = \bar{\mathbf x}_{k-1}, \\ 
  \widehat{\mathbf P}_{i+1} &= \mathbf F_{\widetilde{\mathbf x}_i}\widehat{\mathbf P}_{i}\mathbf F_{\widetilde{\mathbf x}_i}^T + \mathbf F_{\mathbf w_i} \mathbf Q_i \mathbf F_{\mathbf w_i}^T, \ \widehat{\mathbf P}_{0} = \bar{\mathbf P}_{k-1}, 
  \label{eq:state prediction}
\end{aligned} 
\end{equation}
where $\Delta t$ denotes the IMU sampling period, $\mathbf{F}_{\widetilde{\mathbf{x}}}$ and $\mathbf{F}_{\mathbf{w}}$ are the Jacobian of $\mathbf f \left(\mathbf x, \mathbf u, \mathbf w\right)$ w.r.t. $\widetilde{\mathbf x}_i$ and \acrshort{IMU} noise $\mathbf{w_i}$ respectively as detailed in \cite{xu2021fast, xu2022fast}. Notably, $\widehat{\mathbf{Q}}_{k, i}$ denotes the covariance of $\mathbf w_i$, and it is estimated in $(\ref{eq:pred cov estimation})$.

\subsection{State Update}
The \acrshort{MAP} problem modelled in Section \ref{sec:problem definition} could be solved by \acrshort{iEKF} as follows:
\begin{equation} 
\begin{aligned} 
  \begin{split}\;\mathbf{P} & = \left(\mathbf{J}^\kappa\right)^{-1} \widehat{\mathbf{P}}_k\left(\mathbf{J}^\kappa\right)^{-T}, \\
  \mathbf K &= \left( \left( \mathbf H^\kappa \right) ^T {\mathbf R}^{-1} \mathbf H ^\kappa+ {\mathbf P}^{-1} \right)^{-1} \left( \mathbf H^\kappa \right) ^T \mathbf R^{-1}, \\ 
  \widehat{\mathbf x}_{k}^{\kappa +1} & = \widehat{\mathbf x}_{k}^{\kappa } \! \boxplus \! \left(-\mathbf K {\mathbf z}_k^\kappa - (\mathbf I - \mathbf K \mathbf H ^\kappa) (\mathbf J^\kappa)^{-1} \left(\widehat{\mathbf x}_{k}^{\kappa } \boxminus \widehat{\mathbf x}_{k} \right) \right), \end{split} 
\label{eq:state update}
\end{aligned} 
\end{equation}
where $\mathbf J^\kappa$ corresponds to the Jacobian of $\left(\widehat{\mathbf x}_{k}^\kappa \boxplus \widetilde{\mathbf x}_{k}^\kappa \right) \boxminus \widehat{\mathbf x}_k$ w.r.t. $\widetilde{\mathbf x}_{k}^\kappa$, $\mathbf K$ means the Kalman gain, $\mathbf H^\kappa$ is the stacked form of $\mathbf H_{k,j}^\kappa$, and $\mathbf R$ uses the stacked $\mathbf R_{k,j}$ as its diagonal elements. After convergence, the final state $\bar{\mathbf x}_{k}$ and its covariance $\bar{\mathbf P}_{k}$ are updated as:
\begin{equation}
  \begin{split}\; 
    \bar{\mathbf x}_{k} &= \widehat{\mathbf x}_{k}^{\kappa +1},\ \bar{\mathbf P}_{k} = \left(\mathbf I - \mathbf K \mathbf H ^\kappa \right) { \mathbf P }. 
  \end{split} 
\end{equation}

\subsection{Prediction / Measurement Covariance Estimation}
\label{sec:covariance estimation}

After state update, the prediction covariance $\bar{\mathbf{Q}}_{k}$ and the measurement covariance $\bar{\mathbf{R}}_{k,j}$ are updated, and they will be used in state prediction (\ref{eq:state prediction}) and state update (\ref{eq:state update}) of $k+1$th LiDAR scan. 
Based on the deduction in \cite{mohamed1999adaptive,akhlaghi2017adaptive}, $\bar{\mathbf{Q}}_{k}$ is estimated as follows:
\begin{equation}
\begin{aligned}
\mathbf{F}_{{\mathbf{w}},k}&= \sum \limits _{i=1}^{l} \mathbf F_{\mathbf w_i}, \\
\delta {\mathbf{x}_k} &= \bar{\mathbf{x}}_k\boxminus \widehat{\mathbf x}_k,\\
\bar{\mathbf{Q}}_{k} &=  \left( \mathbf{F}_{{\mathbf{w}},k} \right) ^{-1}(\delta {\mathbf{x}_k} \delta {\mathbf{x}_k}^T)\left( \mathbf{F}_{{\mathbf{w}},k} \right) ^{-T}, 
\end{aligned}
\end{equation}
where $l$ is the number of IMU measurements from $k-1$th LiDAR scan end to $k$th LiDAR scan end. 
The key insight of this estimation stems from the relationship between \acrshort{IMU} noise and the state estimation error $\delta {\mathbf{x}_k}$. This error arises during the state prediction phase and is corrected through the subsequent state update process. To quantify \acrshort{IMU} noise covariance, $\delta {\mathbf{x}_k} \delta {\mathbf{x}_k}^T$ is mapped back to the \acrshort{IMU} noise space parameterized by $\mathbf w$ via $\left( \mathbf{F}_{{\mathbf{w}},k} \right) ^{-1}$. In practice, since $\left( \mathbf{F}_{{\mathbf{w}},k} \right) ^{-1}$ is non-diagonal, we compute its blockwise inversion by isolating the contributions of $\mathbf n_{\bm \omega }$, $\mathbf n_{\mathbf a}$, $\mathbf n_{\mathbf b\bm \omega }$ and $\mathbf n_{\mathbf b\mathbf a}$.
Unlike prior works \cite{mohamed1999adaptive,akhlaghi2017adaptive} which assume ${\mathbf{F}_{{\mathbf{w}},k}}$ as an identity matrix, our method employs the accumulated ${\mathbf{F}_{{\mathbf{w}},k}}$ to preserve accuracy.
During \acrshort{IMU} propagation, $\widehat{\mathbf{Q}}_{k}$ keeps unchanged for each \acrshort{IMU} message. To fuse with historical information $\widehat{\mathbf{Q}}_{k}$ for consistency, a forgetting factor $a$ is utilized to update $\widehat{\mathbf{Q}}_{k+1}$:
\begin{equation}
\widehat{\mathbf{Q}}_{k+1}=a \widehat{\mathbf{Q}}_{k} + (1-a)\bar{\mathbf{Q}}_{k}.
\label{eq:pred cov estimation}
\end{equation}

Measurement covariance $\bar{\mathbf{R}}_{k,j}$ consists of two terms. One is the residual $\bar{\mathbf{z}}_j$ which means the error of this constraint that cannot be mitigated after state update, the other is the posterior state covariance $\bar{\mathbf P}_{k}$ projected to the constraint space via $\mathbf H_j^\kappa$:
\begin{equation}
  \bar{\mathbf{R}}_{k,j}=(\bar{\mathbf{z}}_j \bar{\mathbf{z}}_j^T+ \mathbf H_j^\kappa \bar{\mathbf P}_{k} \left( \mathbf {H_j^\kappa} \right)^T).
\label{eq:meas cov estimation}
\end{equation}
Due to the non-repetitive scanning patterns of LiDAR, maintaing $\widehat{\mathbf{R}}_{k+1,j}$ for the LiDAR point ${}^W\bar{\mathbf{p}}_{k,j}$ across successive scans is challenging. Therefore, we update the measurement covariance of the map element associated with ${}^W\bar{\mathbf{p}}_{k,j}$ using $\bar{\mathbf{R}}_{k,j}$ as formalized in (\ref{eq:meas cov update}), and the resulting measurement covariance is utilized to compute $\widehat{\mathbf R}_{k+1,j}$ in (\ref{eq:r inv}).

$\widehat{\mathbf{Q}}_{k+1}$ is used for state prediction in the $k+1$th LiDAR scan to adjust the state estimator's confidence between \acrshort{IMU} and LiDAR measurements. 
For instance, in scenarios where LiDAR degeneracy occurs during the $k$th scan, $\widehat{\mathbf{Q}}_{k+1}$ is reduced accordingly. This adjustment prioritizes \acrshort{IMU} measurements over LiDAR data in the $k+1$th scan to prevent ill-conditioned state estimation, as $\delta{\mathbf{x}}_k$ becomes negligible along the degenerated direction.
During the state update phase of the $k+1$th scan, $\widehat{\mathbf{R}}_{k+1,j}$ quantifies the reliability of the point-to-plane constraint. This reliability stems from the geometric thickness of the associated planar structure and the historical usage of this plane in prior registration processes. If a map element corresponds to a dynamic object, its $\widehat{\mathbf{R}}_{k+1,j}$ increases due to the growth of $\bar{\mathbf{z}}_j$, thereby proportionally reducing its influence on future state estimation.

\section{Voxel-based Gaussian Map}
A voxel-based Gaussian map is proposed to represent planar structures which is constructed by dividing the 3D space into fixed-size voxels (e.g., \SI{0.5}{\meter}) with each voxel containing a set of Gaussian distributions. The map is updated by integrating the LiDAR scan into the Gaussian map via incremental update, and the registration is conducted by minimizing the point-to-plane residual between the LiDAR scan and the Gaussian map with a pseudo-merge correspondence matching strategy.

\subsection{Gaussian Map Representation}
\label{sec:g representation}
The $m$th Gaussian map primitive ${}^W\mathbf{G}_m$ is parameterized by its mean ${}^W\boldsymbol{\mu}_m$, covariance ${}^W\boldsymbol{\Sigma}_m$, observation count $n_m$, the times it has been used in previous registration process $c_m$, and its measurement covariance $\widehat{\mathbf{R}}_{m}$. The $j$th LiDAR point in $k$th scan ${}^{W} \bar{\mathbf p}_{k,j}$ is initialized as an isotropic Gaussian ${}^W{\bar{\mathbf{G}}}_j$ with a fixed covariance in registration and map update.

\subsection{Map Update}
Using the updated state $\bar{\mathbf x}_{k}$, ${}^W\bar{\mathbf{G}}_j$ is merged with the nearest Gaussian ${}^W\mathbf{G}_m$ in the voxel it belongs to based on the Mahalanobis distance $d_{j,m}$ between ${}^W\bar{\mathbf{G}}_j$ and ${}^W\mathbf{G}_m$:
\begin{equation}
\begin{aligned}
  e_{j,m} &= { }^W \bar{\boldsymbol{\mu}}_j-{ }^W \boldsymbol{\mu}_m, \\
  d_{j,m} &= \left(e_{j,m}\right)^T \left({ }^W \bar{\boldsymbol{\Sigma}}_{j}+{}^W\boldsymbol{\Sigma}_m\right)^{-1}{ }^W e_{j,m}.
\end{aligned}
\label{eq:m distance}
\end{equation}
If $d_{j,m}$ satisfies $\mathcal{X}^2$-test with $95 \%$ confidence, ${}^W\mathbf{G}_m$ is fused with ${}^W\bar{\mathbf{G}}_{j}$ using incremental update strategy:
\begin{equation}
  \begin{aligned} 
    r_j &= \frac{n_j}{n_j + n_m},\ r_m = \frac{n_m}{n_j + n_m}, \\
    {}^W\boldsymbol{\mu}_{jm} &= r_j { }^W \bar{\boldsymbol{\mu}}_j + r_m {}^W\boldsymbol{\mu}_m, \\
    {}^W\boldsymbol{\Sigma}_{m} &= r_j \left( { }^W \bar{\boldsymbol{\Sigma}}_j + { }^W \bar{\boldsymbol{\mu}}_j \left( {}^W \bar{\boldsymbol{\mu}}_j \right) {}^T \right) + \\
    &r_m \left( {}^W\boldsymbol{\Sigma}_m + {}^W\boldsymbol{\mu}_m \left( {}^W\boldsymbol{\mu}_m \right) {}^T \right) - {}^W\boldsymbol{\mu}_{jm} \left( {}^W\boldsymbol{\mu}_{jm} \right) {}^T, \\
    {}^W\boldsymbol{\mu}_{m} &= {}^W\boldsymbol{\mu}_{jm}, \\
    n_{m} &= n_j + n_m.
  \end{aligned}
  \label{eq:merge}
\end{equation}
Conversely, if no map correspondence is found, ${}^W\bar{\mathbf{G}}_j$ is added to the voxel as a new Gaussian. 

Update of $\widehat{\mathbf{R}}_{m}$ in this merging process also follows the incremental update strategy by counting $c_m$:
\begin{equation}
  \begin{aligned}
    \widehat{\mathbf{R}}_{m} &= \frac{c_m}{c_m+1}\widehat{\mathbf{R}}_{m} + \frac{1}{c_m+1} \bar{\mathbf{R}}_{k,j}, \\
  c_m &= c_m + 1,
\end{aligned}
\label{eq:meas cov update}
\end{equation}
where $\bar{\mathbf{R}}_{k,j}$ is computed in (\ref{eq:meas cov estimation}).

\subsection{Registration}
\label{sec:registration}
\begin{figure}[t]
  \centering

  \subfloat[Point-to-plane]{\includegraphics[width=0.32\linewidth]{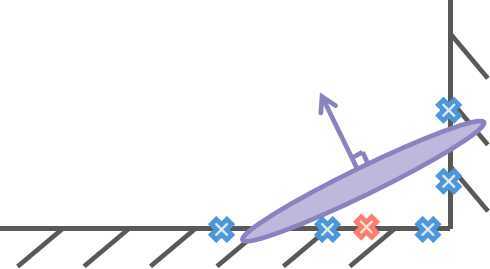}
  \label{fig::reg_a}}
  \subfloat[VGICP]{\includegraphics[width=0.32\linewidth]{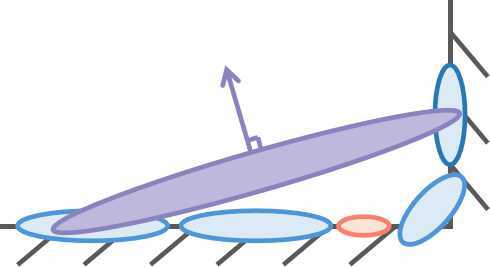}
  \label{fig::reg_b}}
  \subfloat[Pseudo-merge (ours)]{\includegraphics[width=0.32\linewidth]{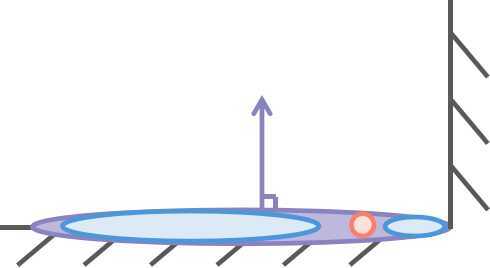}
  \label{fig::reg_c}}
 
  \caption{Different registration strategies and their plane normal estimation near the wall corner (cross: point, ellipse: Gaussian, orange: point/Gaussian in current scan, blue: point/Gaussian in the map, purple: plane for registration).
    (a) Point to k-nearest neighbor points \cite{xu2022fast}. (b) Distribution to k-nearest neighbor distributions \cite{koide2021voxelized, ji2024lio, chen2024ig}. (c) Distribution to k-nearest neighbor distributions with adaptive early termination in our method.}
  \label{fig::reg}
  \vspace{-0.2cm}
\end{figure}

\begin{figure}
\begin{algorithm}[H]
\caption{Correspondence Matching}\label{alg:correspondence matching}
\begin{algorithmic}[1]
  \State \textbf{Input}: 27 neighboring voxels $\mathcal{V}$ of ${}^W\widehat{\mathbf{G}}_j^\kappa$
  \State \textbf{Output}: ${}^W\mathbf{G}_{0m}$
  \State \textbf{Initialize}: $\mathcal{G} \gets \emptyset, {}^W\mathbf{G}_{0m} \gets \emptyset$
  \For{$v \in \mathcal{V}$}
    \For{${}^W\mathbf{G}_m \in v$}
      \State $d_{j,m}^\kappa \gets \textsc{GetDistance}({}^W\widehat{\mathbf{G}}_j^\kappa, {}^W\mathbf{G}_m)$
      \State $\mathcal{G} \gets \mathcal{G} \cup \{\left( d_{j,m}^\kappa, {}^W\mathbf{G}_m \right) \}$
    \EndFor
  \EndFor
  \State $\{ \left( d_{j,0}^\kappa, {}^W\mathbf{G}_0 \right), \dots, \left( d_{j,n}^\kappa, {}^W\mathbf{G}_n \right) \} \gets \textsc{Sort}(\mathcal{G})$
  \For{${}^W\mathbf{G}_m \in \{ \left( d_{j,0}^\kappa, {}^W\mathbf{G}_0 \right), \dots, \left( d_{j,n}^\kappa, {}^W\mathbf{G}_n \right) \}$}
    \State ${}^W\mathbf{G}_{0m} \gets \textsc{PseudoMerge}({}^W\mathbf{G}_{0m}, {}^W\mathbf{G}_{m})$
    \If{\textsc{GetDistance}$({}^W\mathbf{G}_{0m}, {}^L\widehat{\mathbf{G}}_j^\kappa) \le d $}
      \textbf{break}
    \EndIf
  \EndFor
\end{algorithmic}
\end{algorithm}
\vspace{-2em}
\end{figure}

In the $\kappa$-th iteration of the state update, the Gaussian in the LiDAR frame represented as ${}^L{\mathbf{G}}_j^\kappa$ is first transformed to the world frame based on current state estimate $\widehat{\mathbf x}_{k}^{\kappa}$:
\begin{equation}
  \begin{aligned}
  { }^W \widehat{\boldsymbol{\mu}}_j^\kappa & ={ }^W \widehat{\mathbf{T}}_{L}^\kappa { }^{L} \boldsymbol{\mu}_j^\kappa, \\
  { }^W \widehat{\boldsymbol{\Sigma}}_j^\kappa & ={ }^W \widehat{\mathbf{R}}_{L}^\kappa{ }^{L} \boldsymbol{\Sigma}_j^\kappa \left( { }^W \widehat{\mathbf{R}}_{L}^\kappa{ } \right) ^T,
  \end{aligned}
\end{equation}
where ${ }^W \widehat{\mathbf{T}}_{L}^\kappa$ and ${ }^W \widehat{\mathbf{R}}_{L}^\kappa$ are the transformation and rotation of LiDAR w.r.t. the world frame.
Under the planarity assumption, the point-to-plane residual $\mathbf{z}_{k, j}^\kappa$ is computed as follows: 
\begin{equation}
  \label{eq:residual}
  \mathbf{z}_{k, j}^\kappa=\mathbf{u}_{0m}^T \left({ }^W \widehat{\boldsymbol{\mu}}_j^\kappa-{ }^W \boldsymbol{\mu}_{0m}\right),
\end{equation}
where ${}^W\mathbf{G}_{0m}$ denotes the map correspondence of ${}^W\widehat{\mathbf{G}}_j^\kappa$ generated via the pseudo-merge strategy (Algorithm \ref{alg:correspondence matching}), and $\mathbf{u}_{0m}$ is the normal vector of ${}^W\mathbf{G}_{0m}$. The uncertainty $\widehat{\mathbf R}_{k,j}$ associated with this constraint is computed as follows:
\begin{equation}
  \begin{aligned}
    c_{0m} &= \sum \limits _{l=0}^{m}c_l, \\
  \mathbf R_{0m} &= \sum \limits _{l=0}^{m} \frac{c_l}{c_{0m}}\widehat{\mathbf{R}}_{l}, \\
  s_{0m} &= \mathbf{n}_{0m}^T {}^W\boldsymbol{\Sigma}_{0m} \mathbf{n}_{0m}, \\ 
  \widehat{\mathbf R}_{k,j} &= e^{b\mathbf R_{0m}} s_{0m},
\end{aligned}
  \label{eq:r inv}
\end{equation}
where $s_{0m}$ quantifies the squared geometric thickness of the planar structure ${}^W\mathbf{G}_{0m}$, $\mathbf R_{0m}$ corresponds to the historical uncertainty of ${}^W\mathbf{G}_{0m}$ being used as constraints in previous state updates, and $b$ is a positive scaling factor amplifying influence of $\mathbf R_{0m}$ on $\widehat{\mathbf R}_{k,j}$ within the exponential term. By jointly leveraging the historical reliability and current geometric fidelity of $\mathbf R_{0m}$, this formulation enhances the uncertainty of the constraints associated with dynamic objects. For example, if ${}^W\mathbf{G}_{0m}$ corresponds to a moving object, $s_{0m}$ may be small due to LiDAR point sparsity, yet $\widehat{\mathbf R}_{k,j}$ remains large if historical residuals of ${}^W\mathbf{G}_{0m}$ are persistently large as computed in (\ref{eq:meas cov estimation}).

The correspondence matching pipeline is formalized in Algorithm \ref{alg:correspondence matching} and illustrated in Fig. \ref{fig::framework}. The search space is constrained to the 27 neighboring voxels of ${}^W\widehat{\mathbf{G}}_j^\kappa$, within which the Mahalanobis distance $d_{j,m}^\kappa$ between ${}^W\mathbf{G}_m$ and ${}^W\widehat{\mathbf{G}}_j^\kappa$ is computed as in (\ref{eq:m distance}). All the map Gaussians in these voxels are sorted based on $d_{j,m}^\kappa$ in increasing order, and we employ a pseudo-merge strategy (\ref{eq:merge}) to ensure ${}^W\widehat{\mathbf{G}}_j^\kappa$ resides within the high-confidence region of its correspondence ${}^W\mathbf{G}_{0m}$ using this sorted list. Starting with the closest Gaussian ${}^W\mathbf{G}_0$ in the sorted list, it iteratively merges subsequent Gaussians until the Mahalanobis distance between ${}^W\mathbf{G}_{0m}$ and ${}^W\widehat{\mathbf{G}}_j^\kappa$ falls below a predefined threshold $d$. Once ${}^W\mathbf{G}_{0m}$ satisfies this criterion, the residual for this constraint is derived as in (\ref{eq:residual}). 

If ${}^W\mathbf{G}_m$ is pseudo-merged with ${}^W\mathbf{G}_{0m}$ during registration, its measurement covariance $\widehat{\mathbf{R}}_{m}$ is updated accordingly after the state update as computed in (\ref{eq:meas cov update}) and illustrated at the bottom left of Fig. \ref{fig::framework}.

As shown in Fig. \ref{fig::reg_c}, the proposed pseudo-merge strategy achieves precise plane normal estimation, particularly in non-planar regions like wall corners, owing to its adaptive early termination criterion. In contrast, FAST-LIO2 \cite{xu2022fast} suffers from inaccuracies in normal estimation due to map point sparsity (Fig. \ref{fig::reg_a}), while voxelized Gaussian-based methods \cite{koide2021voxelized, ji2024lio, chen2024ig} exhibit errors caused by over-smoothed plane merging (Fig. \ref{fig::reg_b}). Notably, inaccurate plane normals degrade state estimation robustness, particularly under LiDAR degeneracy, where estimation stability is highly sensitive to noise along the unobservable direction.

%% file: sections/experiment.tex
\section{Experiments}
\label{sec:exp}
We conduct comprehensive evaluations of AKF-LIO across multiple benchmark datasets, including the MARS-LVIG Dataset \cite{li2024mars}, the KITTI Odometry Benchmark \cite{geiger2013vision} and the R3LIVE Dataset \cite{lin2022r}. Moreover, extensive ablation studies and performance analysis reveal the effectiveness of the proposed modules. All experiments are conducted on a desktop computer with an Intel-i9-14900KF CPU (24 cores @ 3.2 GHz) and 64GB RAM.

\subsection{MARS-LVIG Dataset}
\begin{table*}[t]
    \setlength\tabcolsep{3.8pt} 
    \caption{RMSE of ATE (METERS) RESULTS ON MARS-LVIG DATASET}
    \centering
    \begin{tabularx}{\textwidth}{@{}l ccc ccc ccc ccc ccc ccc ccc ccc@{}}
      \toprule
        &  & \multicolumn{3}{c}{HKairport} & \multicolumn{3}{c}{HKairport\_GNSS}
        & \multicolumn{3}{c}{HKisland} & \multicolumn{3}{c}{HKisland\_GNSS}
        & \multicolumn{3}{c}{AMtown} & \multicolumn{3}{c}{AMvalley} &
        \multicolumn{3}{c}{Featureless\_GNSS}\\
      
        \cmidrule(lr){3-5} \cmidrule(lr){6-8} \cmidrule(lr){9-11} \cmidrule(lr){12-14} 
        \cmidrule(lr){15-17} \cmidrule(lr){18-20} \cmidrule(lr){21-23}
        & & \multicolumn{1}{c}{01} & \multicolumn{1}{c}{02} & \multicolumn{1}{c}{03} &
        \multicolumn{1}{c}{01} & \multicolumn{1}{c}{02} & \multicolumn{1}{c}{03} &
        \multicolumn{1}{c}{01} & \multicolumn{1}{c}{02} & \multicolumn{1}{c}{03} &
        \multicolumn{1}{c}{01} & \multicolumn{1}{c}{02} & \multicolumn{1}{c}{03} &
        \multicolumn{1}{c}{01} & \multicolumn{1}{c}{02} & \multicolumn{1}{c}{03} &
        \multicolumn{1}{c}{01} & \multicolumn{1}{c}{02} & \multicolumn{1}{c}{03} &
        \multicolumn{1}{c}{01} & \multicolumn{1}{c}{02} & \multicolumn{1}{c}{03} \\
  
      \midrule
      FAST-LIO2 & & 0.44 & 0.96 &  1.28 & 0.52 & \textbf{2.92} & 0.62 & 0.66 & 1.56 & 2.13 & 1.88 & 
      2.07 & 1.75 & 3.18 & 3.53 & 3.65 & 5.19 & 8.14 & 7.77 & - & 12.03 & -\\
      iG-LIO & & - & - &  - & - & - & - & \textbf{0.28} & - & - & \textbf{1.83} & 
      \textbf{2.03} & - & 1.65 & - & 3.54 & - & - & - & - & 3.45 & -\\
      R3LIVE & & 0.68 & \textbf{0.82} & \textbf{1.12} & 0.53 & 2.98 & 1.41 & 0.70 & 2.10 & 3.93 & 2.02 & 2.11 & 
      3.68 & 1.44 & \textbf{2.10} & - & 3.82 & 4.47 & - & - & - & -\\ 
      Ours & & \textbf{0.43} & 0.87 & 1.23 & \textbf{0.46} & 2.94 & \textbf{0.59} & \textbf{0.28} 
      & \textbf{1.46} & \textbf{2.03} &1.84 & 2.04 & \textbf{1.52} & \textbf{1.25} & 2.21 & \textbf{2.74} 
      & \textbf{2.57} & \textbf{4.06} & \textbf{3.91} & \textbf{1.91} & \textbf{3.23} & \textbf{4.74}\\

      \bottomrule
    \end{tabularx}
  \vspace{-1.0em} 
  \label{table:MARS}
\end{table*}

\begin{figure}[t]
  \centering
  \includegraphics[width=3.4in]{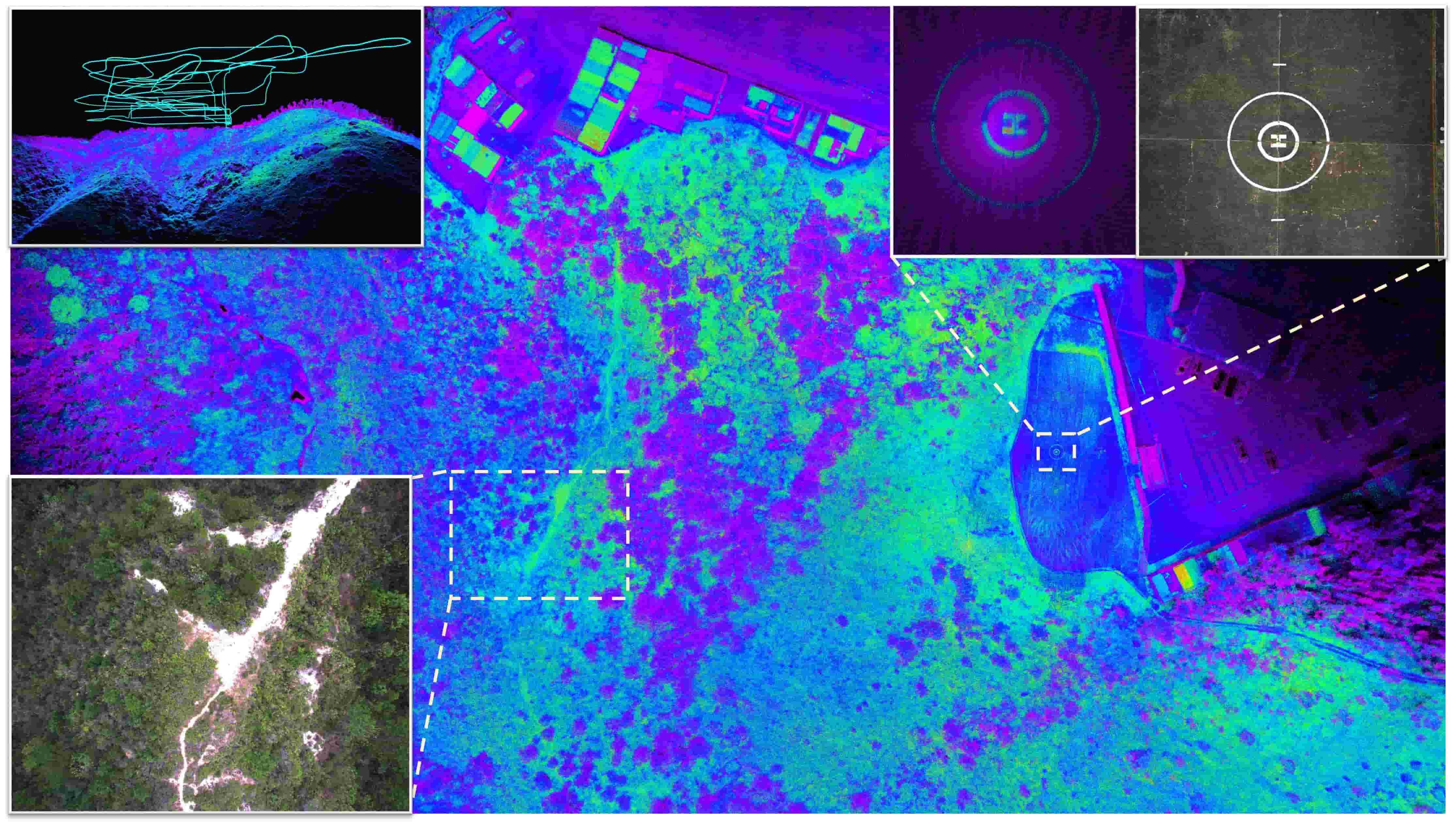}
  \vspace{-1.5em}
  \caption{Accumulated LiDAR point clouds (colored by LiDAR intensity) of Featureless\_GNSS03 sequence in MARS-LVIG dataset \cite{li2024mars}. The top-left corner shows the estimated trajectory of the unmanned aerial vehicle platform used for data collection. The bottom-left depicts the camera view of the unstructureless forest environment. The top-right shows the LiDAR and camera views at the taking-off and landing position where geometry degraded because of single plane exists in the LiDAR FoV.}
  \label{fig::mars03}
  \vspace{-1.5em}
\end{figure}
The MARS-LVIG dataset \cite{li2024mars}, a comprehensive multi-sensor aerial SLAM benchmark, is used to evaluate our method against FAST-LIO2 \cite{xu2022fast}, iG-LIO \cite{chen2024ig}, and R3LIVE \cite{lin2022r}. FAST-LIO2 leverages a filter-based approach, fusing LiDAR and \acrshort{IMU} measurements via an \acrshort{iEKF} with ikd-Tree for registration. iG-LIO adopts incremental Generalized Iterative Closest Point (GICP) with surface covariance estimation akin to our method. R3LIVE integrates LiDAR, \acrshort{IMU} and camera, employing separate subsystems for geometric mapping via \acrshort{LIO} and texture mapping via \acrfull{VIO}. All methods utilize a Livox Avia LiDAR and \acrshort{IMU}, while R3LIVE incorporates an additional RGB camera.

As outlined in Table \ref{table:MARS}, FAST-LIO2 and R3LIVE achieve consistent performance across the majority of the sequences. For R3LIVE, official results are reported from the MARS-LVIG Dataset \cite{li2024mars} due to irreproducibility with its default parameters. 
In the Featureless\_GNSS sequences, manual flights conducted at a low altitude of \SI{20}{\meter} lead to LiDAR and visual degradations due to the sensor's \acrshort{FoV} being dominated by flat and textureless surfaces. Our method maintains reliable localization compared to other methods in Featureless\_GNSS sequences by adaptively weighting sensor confidence via noise covariance estimation when individual sensors degrade. Our framework outperforms competitors in the AMvalley forest sequences, where FAST-LIO2 and R3LIVE exhibit over-reliance on LiDAR data, despite noise from dense tree leaves. This robustness arises from our Gaussian map representation, which inherently accommodates sensor noise through probabilistic modeling. While iG-LIO achieves competitive accuracy in stable scenarios, it fails in most sequences, likely due to its single-Gaussian-per-voxel limitation as described in Section \ref{sec:gaussian related work}, which inadequately models complex geometries under noisy LiDAR observations. Overall, our method demonstrates consistent robustness, achieving first or second accuracy across all sequences. The qualitative result of the Featureless\_GNSS03 sequence is visualized in Fig.\ref{fig::mars03}.

\subsection{KITTI Odometry Benchmark}

\begin{table}[t]
  \renewcommand{\arraystretch}{1.6}
    \setlength\tabcolsep{1.3pt} 
    \caption{RTE RESULTS (\%) ON KITTI ODOMETRY BENCHMARK}
    \centering
    \begin{tabularx}{\columnwidth}{c ccccccccccc c}
    \toprule
        & 00 & 01 & 02 & 03 & 04 & 05 & 06 & 07 & 08 & 09 & 10 & Online \\
      \midrule
      KISS-ICP & 0.52 & 0.72 & 0.53 & \textbf{0.65} & \textbf{0.35} & 0.30 & \textbf{0.26} & 0.33 & 0.81 & 0.49 & 0.54 & 0.61 \\
      CT-ICP & \textbf{0.49} & 0.76 & 0.52 & 0.72 & 0.39 & \textbf{0.25} & 0.27 & 0.31 & 0.81 & 0.49 & \textbf{0.48} & 0.59 \\
      Traj-LO & 0.50 & 0.81 & 0.52 & 0.67 & 0.40 & \textbf{0.25} & 0.27 & \textbf{0.30} & 0.81 & \textbf{0.45} & 0.55 & \textbf{0.58} \\
      VoxelMap & 0.82 & 0.85 & 1.71 & 0.69 & 0.44 & 0.40 & 0.35 & 0.35 & 0.88 & 0.50 & 0.66 & / \\ 
      VoxelMap-L & 0.59 & 0.86 & 0.67 & 0.69 & 0.44 & 0.32 & 0.33 & 0.34 & \textbf{0.79} & 0.50 & 0.63 & / \\
      Ours & \textbf{0.49} & \textbf{0.65} & \textbf{0.50} & 0.66 & 0.37 & 0.26 & 0.29 & 0.32
      & 0.80 & 0.46 & 0.54 & 0.59\\
    \bottomrule
  \end{tabularx}
  \vspace{-1.5em} 
  \label{table:KITTI}
\end{table}

\begin{figure}[t]
  \vspace{0.5em}
  \centering
  \includegraphics[width=3.4in]{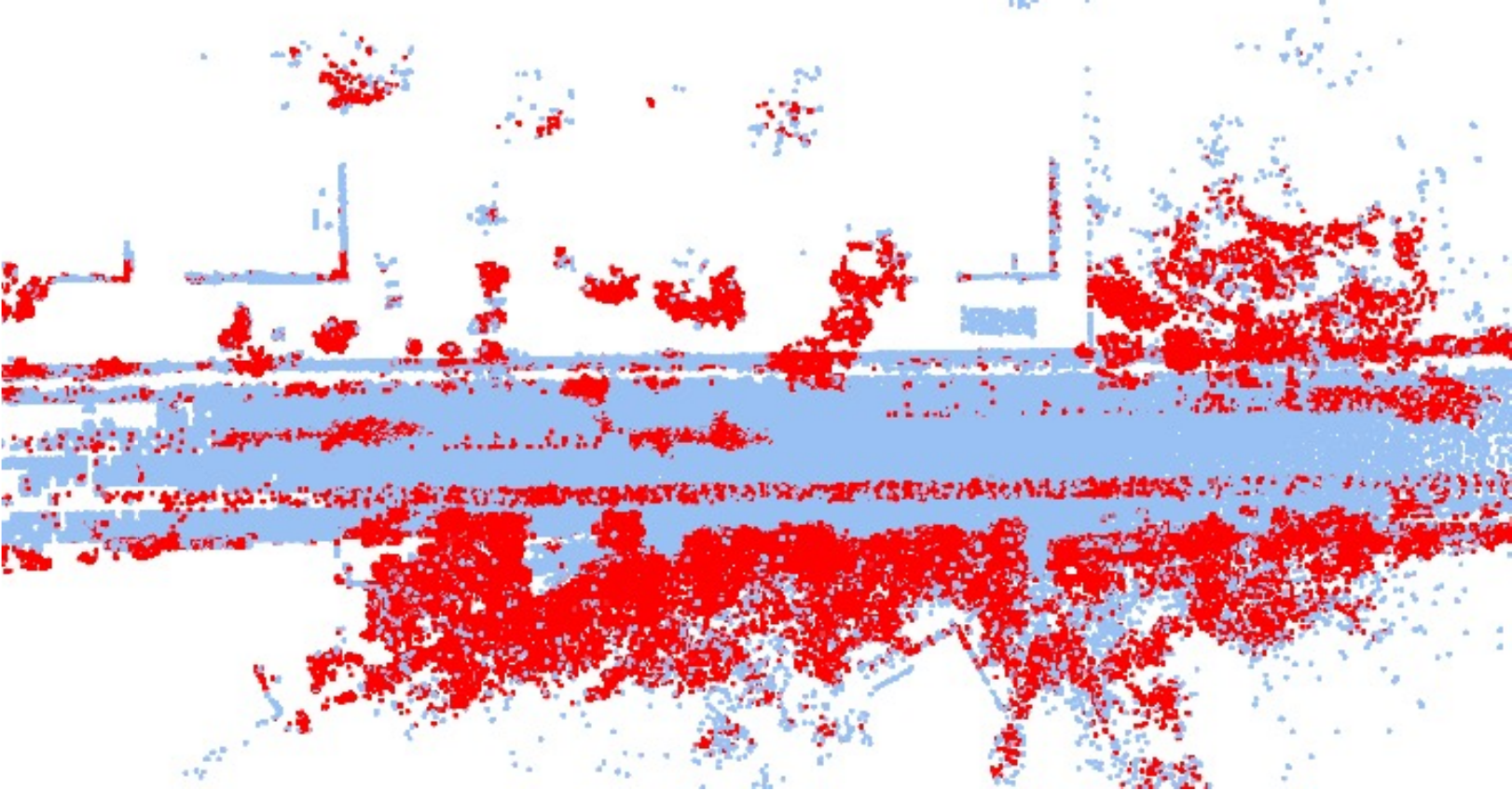}
  \vspace{-1.5em}
  \caption{Visualization of the measurement covariance values stored in the Gaussian map on the KITTI 04 sequence. Large measurement covariance (red) corresponds to dynamic objects like vehicles and nonplanar shapes like corners and tree leaves, and small measurement covariance (blue) corresponds to static planar surfaces like buildings and roads.}
  \label{fig::kitti}
  \vspace{-1.5em}
\end{figure}

The KITTI Odometry Benchmark \cite{geiger2013vision} as a widely recognized autonomous driving dataset is employed to evaluate our method. Since IMU data is unavailable, we disable the IMU prediction module and adopt a constant velocity model. Our method is compared with LiDAR-only algorithms, including KISS-ICP \cite{vizzo2023kiss} with point-to-point registration using adaptive correspondence thresholds, CT-ICP \cite{dellenbach2022ct} with continuous-time intra-scan and inter-scan modeling, Traj-LO \cite{zheng2024traj} with high-frequency estimation via scan segmentation, and VoxelMap \cite{yuan2022efficient} with octree-based voxel management with uncertainty-aware planes.

Table \ref{table:KITTI} reports results using \acrfull{RTE} metric. All methods achieve comparable accuracy on sequences 00-10 except VoxelMap, whose global map strategy adversely impacts \acrshort{RTE}. Therefore, we implement VoxelMap-L, a local map variant that discards map elements older than \SI{5}{\second}, to improve its \acrshort{RTE} performance. On online benchmark sequences, Traj-LO leads with 0.58\% \acrshort{RTE}, followed closely by CT-ICP (0.59\%) and our algorithm (0.59\%). The online sequence results of VoxelMap are missed since they are not provided in their work. Our method currently ranks 8th on the KITTI Odometry Benchmark\footnote{\url{https://www.cvlibs.net/datasets/kitti/eval_odometry.php}}.
The superior performance of CT-ICP and Traj-LO potentially stems from their continuous-time trajectory modeling, which aligns with RTE's emphasis on local consistency, as proved by VoxelMap-L's superior performance compared with original VoxelMap. Nevertheless, there are 30 parameters in CT-ICP in their configuration file, and the segment number of Traj-LO varies across different datasets. In contrast, \acrshort{AKF} in our method integrates online covariance estimation with Gaussian map representation, enabling nearly unified parameterization across diverse datasets. As illustrated in Fig. \ref{fig::kitti}, our covariance estimation module assigns larger uncertainties to dynamic objects and non-planar structures while reserving lower uncertainties for static planar surfaces, enhancing robustness in heterogeneous environments.

\subsection{R3LIVE Dataset}
\begin{table}[t]
  \centering
  \setlength\tabcolsep{6pt} 
  \caption{END-TO-END ERRORS (METERS) ON R3LIVE DATASET}
  \begin{tabularx}{\columnwidth}{l ccccccc}
    \toprule
    &  \multicolumn{3}{c}{degenerate\_seq} & \multicolumn{3}{c}{hkust\_campus}\\
        \cmidrule(lr){2-4} \cmidrule(lr){5-7}
        & \multicolumn{1}{c}{00} & \multicolumn{1}{c}{01} & \multicolumn{1}{c}{02} &
        \multicolumn{1}{c}{00} & \multicolumn{1}{c}{01} & \multicolumn{1}{c}{02} \\
  
      \midrule
      Duration(s) & 86 & 85 & 101 & 1073 & 1162 & 478 \\
      Length(m) & 53.3 & 75.2 & 74.9 & 1317.2 & 1524.3 & 503.8 \\
      \midrule
      FAST-LIO2 & 8.64 & 6.68 &  49.06 & 5.20 & \textbf{0.14} & 0.12 \\
      R3LIVE & 0.07 & \textbf{0.09} & 0.10 & 3.70 & 21.61 & 0.06 \\ 
      iG-LIO & 47.41 & 5.35 & - & 2.81 & 3.44 & 0.06 \\
      Ours & \textbf{0.04} & 13.26 & \textbf{0.02} & \textbf{0.05} & 2.07 & \textbf{0.02} \\
    \bottomrule
  \end{tabularx}
  \label{table::r3live}
\vspace{-0.5em} 
\end{table}

We further evaluate our method in the R3LIVE \cite{lin2022r} dataset, as it contains multiple LiDAR degenerated sequences. The groundtruth trajectory is not available and the end-to-end errors are reported in the Table \ref{table::r3live}. All the methods show similar results on the non-degenerated sequences. For the degenerated\_seq 00-02, FAST-LIO2 and iG-LIO quickly drift when only one plane exists in the LiDAR \acrshort{FoV}, like facing the wall or the ground. R3LIVE shows great robustness across the three degenerated sequences by introducing camera measurements. Our method can survive in two of the degenerated sequences owing to \acrshort{AKF} and robust registration with pseudo-merged planes. A possible reason for the failure in degenerate\_seq\_01 may be that the LiDAR degeneracy occurs at the beginning of the sequence where the prediction covariance estimation module fails to converge in such a short time.

\subsection{Ablation Study}

\begin{table}[t]
  \centering
  \setlength\tabcolsep{6pt} 
  \caption{RMSE OF ATE (METERS) RESULTS ON Field-Dynamic Sequence of ENWIDE LiDAR Inertial Dataset. }
  \begin{tabularx}{\columnwidth}{ll ccc}
    \toprule
    Prediction &  Measurement & FAST-LIO2 & Ours-iEKF & Ours-AKF \\
    \midrule
    100$\mathbf{Q}$ & 100$\mathbf{R}$ & 23.78 & \textbf{0.18} & 0.20  \\
    100$\mathbf{Q}$ & 1$\mathbf{R}$ & 14.40 & 0.19 &  \textbf{0.18}  \\
    1$\mathbf{Q}$ & 100$\mathbf{R}$ & 45.80 & 89.46 &  \textbf{0.20} \\
    1$\mathbf{Q}$ & 1$\mathbf{R}$ & 14.17 & \textbf{0.18} &  \textbf{0.18} \\
    0.01$\mathbf{Q}$ & 0.01$\mathbf{R}$ & 11.82 & \textbf{0.18} & 0.19 \\
    0.01$\mathbf{Q}$ & 1$\mathbf{R}$ & 40.35 & 7.51 &  \textbf{0.18} \\
    1$\mathbf{Q}$ & 0.01$\mathbf{R}$ & 0.78 & 0.20 &  \textbf{0.18} \\
    \bottomrule
  \end{tabularx}
\label{table:coin}
\vspace{-1.5em} 
\end{table}

\begin{figure}[t]
  \vspace{0.5cm}
  \centering
  \includegraphics[width=0.32\linewidth]{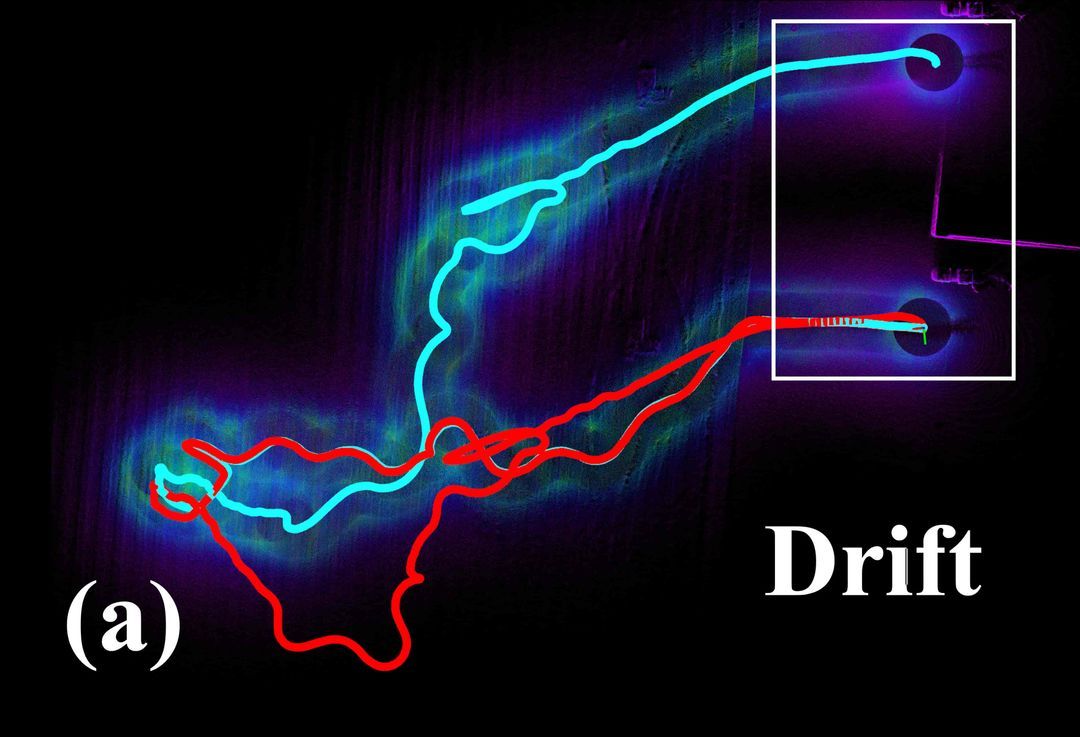}
  \includegraphics[width=0.32\linewidth]{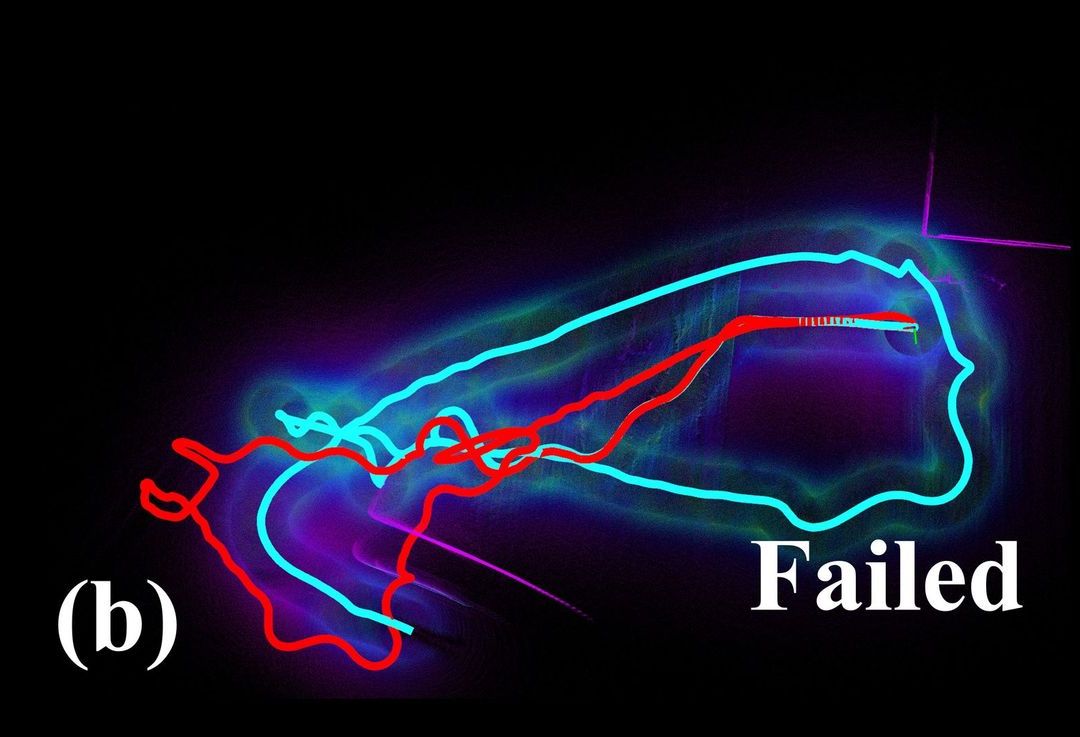}
  \includegraphics[width=0.32\linewidth]{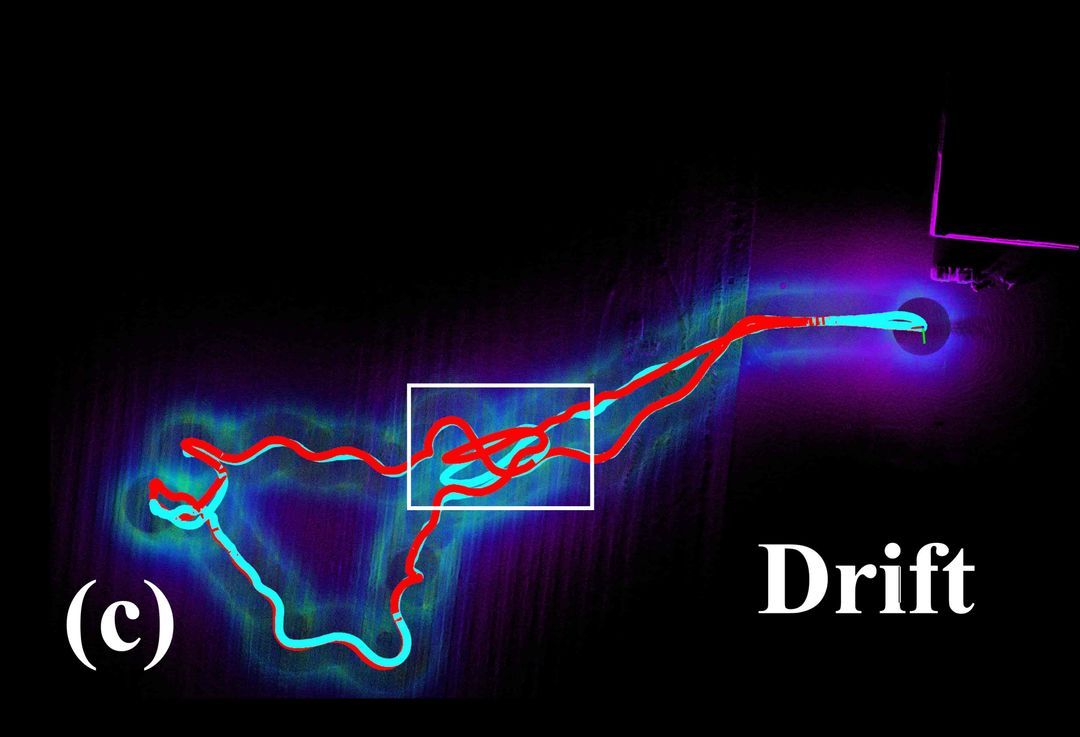}\vspace{0.1cm}\\

  \includegraphics[width=0.32\linewidth]{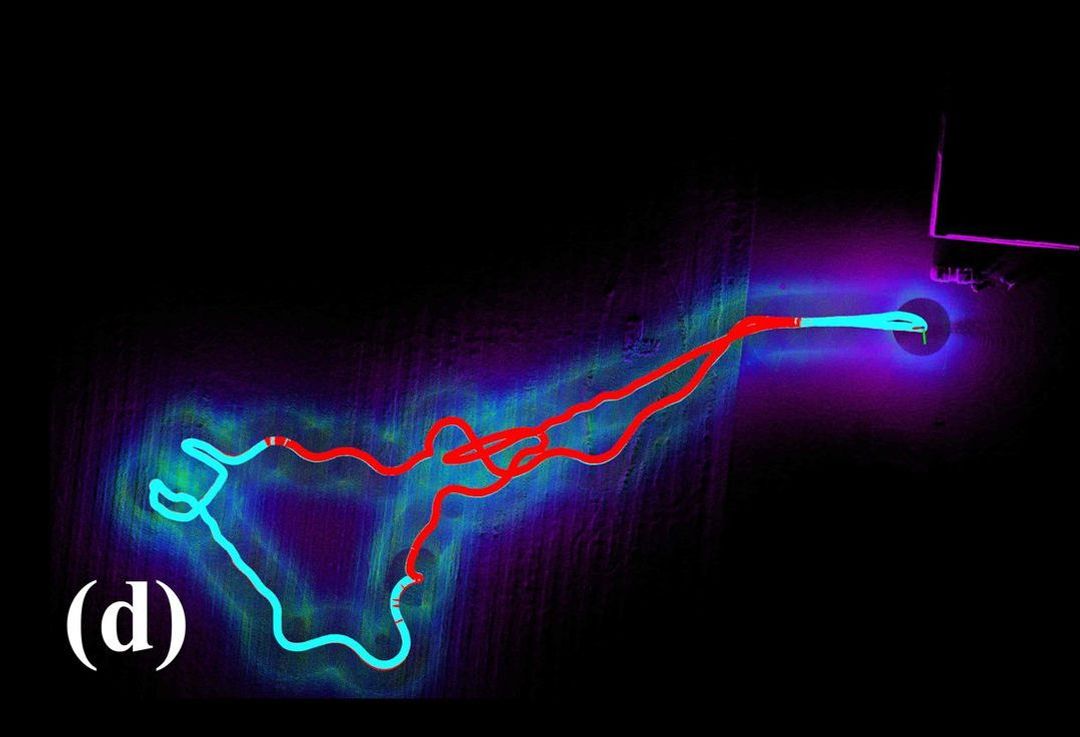}
  \includegraphics[width=0.32\linewidth]{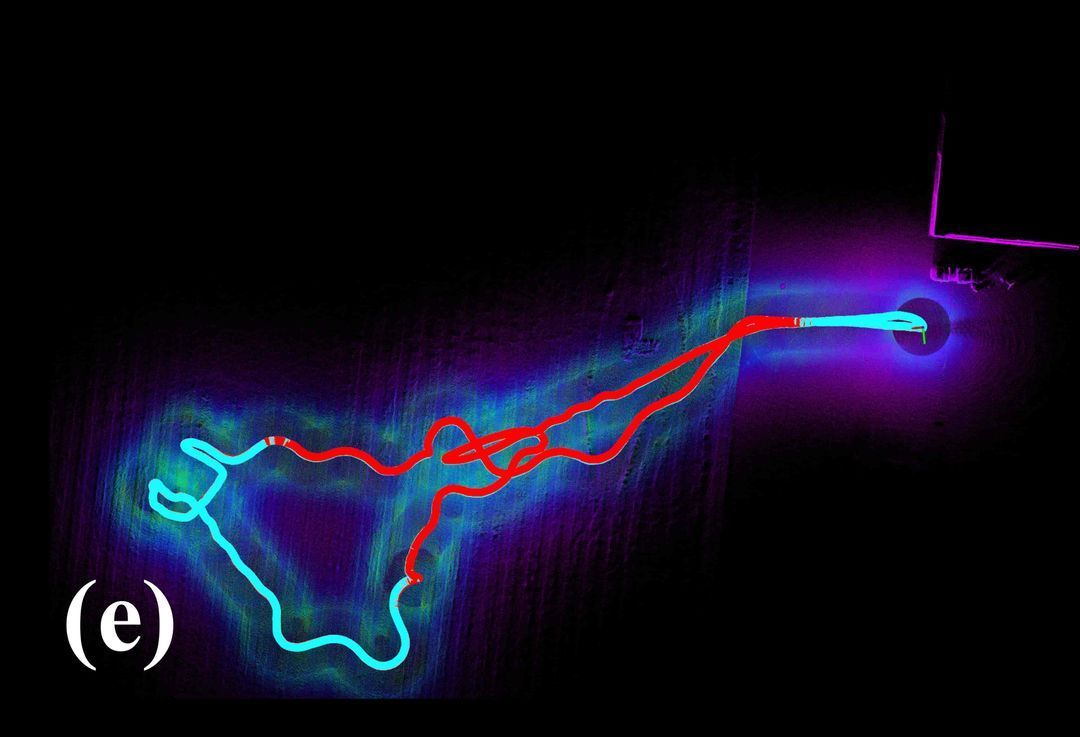}
  \includegraphics[width=0.32\linewidth]{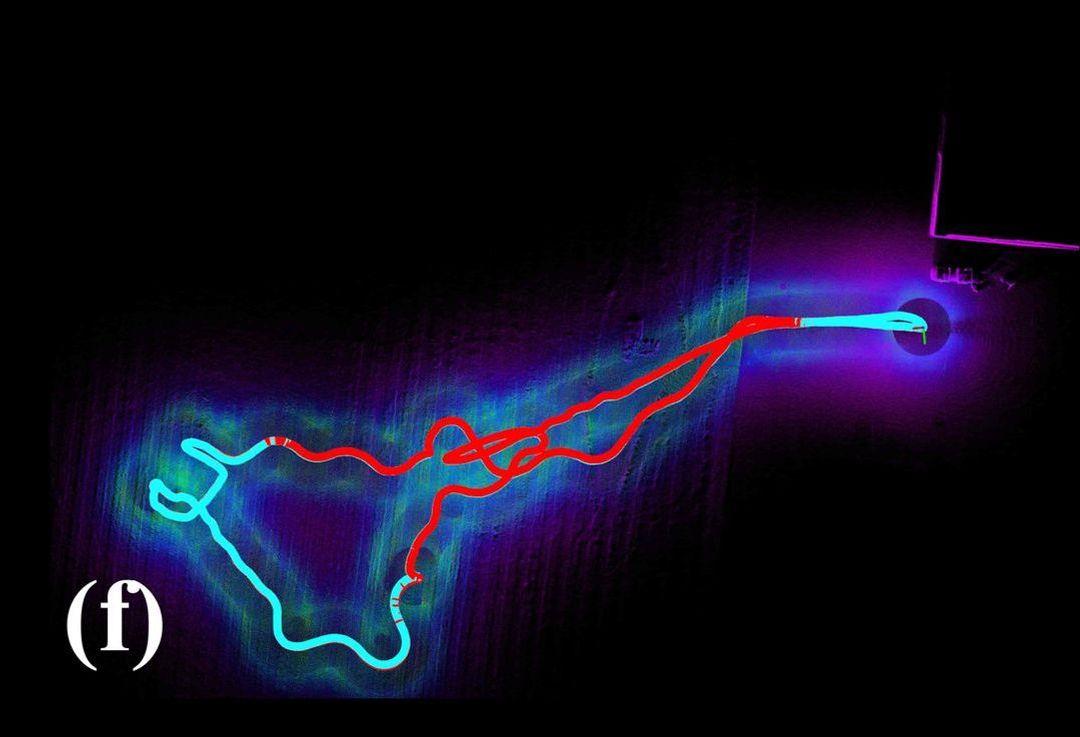}\\
 
  \caption{Qualitative results of FAST-LIO2 (a-c) and ours-AKF (d-f) on Field-Dynamic Sequence of ENWIDE LiDAR Inertial Dataset by setting different initial prediction and measurement covariances:
    \textbf{(a)} FAST-LIO2 (1$\mathbf{Q}$ and 1$\mathbf{R}$),
    \textbf{(b)} FAST-LIO2 (0.01$\mathbf{Q}$ and 1$\mathbf{R}$),
    \textbf{(c)} FAST-LIO2 (1$\mathbf{Q}$ and 0.01$\mathbf{R}$),
    \textbf{(d)} AKF-LIO (1$\mathbf{Q}$ and 1$\mathbf{R}$),
    \textbf{(e)} AKF-LIO (0.01$\mathbf{Q}$ and 1$\mathbf{R}$),
    \textbf{(f)} AKF-LIO (1$\mathbf{Q}$ and 0.01$\mathbf{R}$).
    The red paths show the ground truth trajectories and the cyan paths depict the estimated trajectories. 
    The corresponding RMSE results are reported in Table \ref{table:coin}.}
  \label{fig:ablation}
  \vspace{-1.5em}
\end{figure}

\begin{figure}[t]
  \centering
  \includegraphics[width=3in]{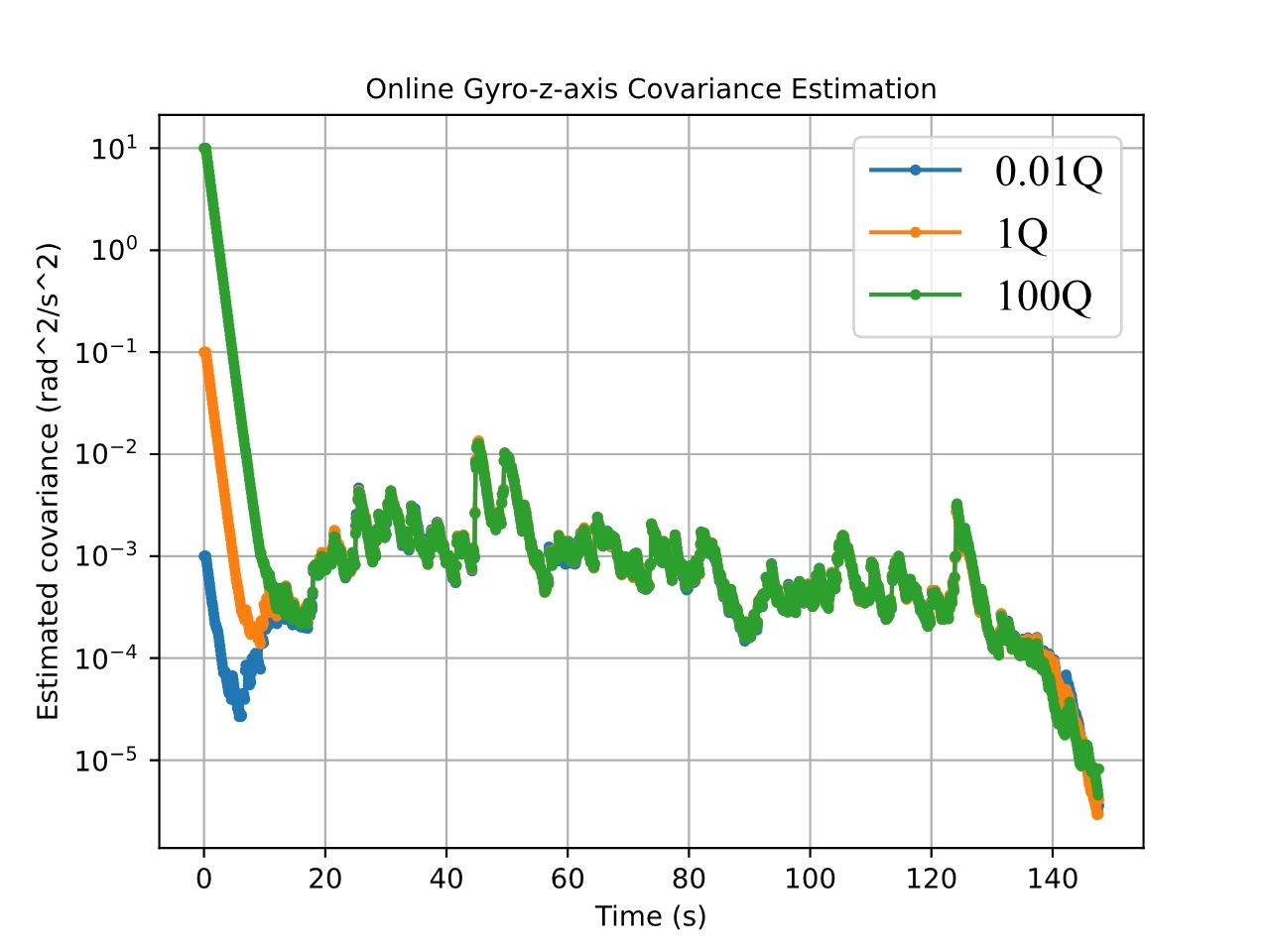}
  \vspace{-0.5em}
  \caption{Online Gyro-z-axis covariance estimation with the same measurement noise (1$\mathbf{R}$) and three different prediction covariance (0.01$\mathbf{Q}$, 1$\mathbf{Q}$, 100$\mathbf{Q}$). The estimated prediction covariances quickly converge to the same level after movement at about 10 seconds.}
  \label{fig::akf_gyro}
\end{figure}

The ablation study is conducted on the field-dynamic sequence of ENWIDE LiDAR Inertial Dataset originally used by COIN-LIO \cite{pfreundschuh2023coin}. It is a short sequence containing the LiDAR-degenerated open environment, which is suitable for validating the robustness of our method in such challenging cases. Three methods are compared in this sequence, including FAST-LIO2, our method utilizing fixed covariance (Ours-iEKF) as in FAST-LIO2 and our original method (Ours-AKF). $\mathbf{Q}$ and $\mathbf{R}$ represent the default fixed prediction and measurement covariance values used in FAST-LIO2 respectively, and $\mathbf{R}$ is utilized to initialize $\widehat{\mathbf{R}}_{m}$ in Ours-AKF method. We initialize the system covariance parameter with 100$\times$, 1$\times$, 0.01$\times$ scaled $\mathbf{Q}$ and $\mathbf{R}$.

\acrfull{RMSE} of \acrfull{ATE} results are reported in the Table \ref{table:coin}. FAST-LIO2 shows overconfidence on the constructed plane and drifts in most of the parameter settings, since it is geometrically under-constrained when only a single ground plane exists. For reference, COIN-LIO gets \SI{0.581}{\meter} RMSE of ATE in this sequence using extra intensity channel measurement of LiDAR. Ours-iEKF, using fixed covariance parameters, performs better than FAST-LIO2 because of the better plane normal estimation of the probabilistic plane constructed by the pseudo-merge strategy. Ours-AKF with online covariance estimation shows the best robustness and achieves similar accuracy across all the varying initial settings. The results demonstrate that \acrshort{AKF} is insensitive to the initial covariance values. Qualitative comparisons between FAST-LIO2 and ours-AKF with different parameter settings are shown in Fig. \ref{fig:ablation}. 

We further visualize the curves of the estimated gyroscope covariance values along its z-axis in Fig. \ref{fig::akf_gyro} to show the quick convergence of \acrshort{AKF} despite different initial values. As demonstrated in Fig. \ref{fig::akf_gyro}, the covariance estimation quickly decreases to a certain level since the robot stays static at the beginning of the sequence, indicating that the system has a high confidence on \acrshort{IMU} in this period. After \SI{10}{\second}, when the operator starts to move, the three estimated noises also quickly converge to the same level, followed by a drop near the sequence end when the operator goes back to the origin and remains static. This means the system has higher confidence on IMU when the operator keeps static as IMU is less noisy in such scenarios.

\subsection{Performance Analysis}

\begin{table}[t]
  \centering
  \setlength\tabcolsep{8pt} 
  \caption{PERFORMANCE EVALUATION ON AMtown-01 Sequence of MARS-LVIG DATASET}
  \begin{tabularx}{\columnwidth}{l cccc}
    \toprule
    &  FAST-LIO2 & iG-LIO & VoxelMap & Ours \\
    \midrule
    Time (ms) & 19.51 & \textbf{18.20} & 48.14 & 19.05 \\
    Memory (GB) & \textbf{2.20} & 6.67 & 14.15 & 4.32  \\
    \bottomrule
  \end{tabularx}
\vspace{-1.5em} 
\label{table:performance}
\end{table}

Time and memory consumption are evaluated to show the computational efficiency of our method as illustrated in Table \ref{table:performance}. As a result, all the methods used for comparison have real-time processing capability. Notably, VoxelMap performs the worst as it stores much more LiDAR points than the other three methods for octree subdivisions. FAST-LIO2 takes the least memory by utilizing a sparse ikd-Tree map representation\cite{cai2021ikd}. Our method shows comparable processing speed and efficient memory usage by utilizing a fine-to-coarse Gaussian representation.